\documentclass{article}
\usepackage[margin=1in]{geometry}
\usepackage[utf8]{inputenc}
\usepackage{amsmath,amssymb,amsthm}
\usepackage{hyperref}
\hypersetup{colorlinks,allcolors=[rgb]{0,0.13672,0.95703}}
\usepackage{cleveref,xcolor}
\usepackage[affil-sl]{authblk}
\usepackage{graphicx}
\usepackage[citestyle=alphabetic,natbib=true,bibstyle=alphabetic,maxbibnames=99]{biblatex}

\title{Two models of double descent for weak features}

\author[1]{Mikhail Belkin}
\author[2,3]{Daniel Hsu}
\author[2]{Ji Xu}
\affil[1]{Halıcıoğlu Data Science Institute, UC San Diego, La Jolla, CA}
\affil[2]{Department of Computer Science, Columbia University, New York, NY}
\affil[3]{Data Science Institue, Columbia University, New York, NY}

\newcommand\T{{*}}

\newcommand\tr{{\operatorname{tr}}}

\newcommand\E{{\mathbb{E}}}
\newcommand\C{{\mathbb{C}}}
\newcommand\R{{\mathbb{R}}}
\newcommand\vP{{\boldsymbol{P}}}
\newcommand\vPi{{\boldsymbol{\Pi}}}
\newcommand\vF{{\boldsymbol{F}}}

\newcommand\vI{{\boldsymbol{I}}}
\newcommand\vK{{\boldsymbol{K}}}
\newcommand\vX{{\boldsymbol{X}}}
\newcommand\vv{{\boldsymbol{v}}}
\newcommand\vx{{\boldsymbol{x}}}
\newcommand\vy{{\boldsymbol{y}}}
\renewcommand\v[1]{{\boldsymbol{#1}}}
\newcommand\vbeta{{\boldsymbol{\beta}}}
\newcommand\vmu{{\boldsymbol{\mu}}}
\newcommand\veta{{\boldsymbol{\eta}}}
\newcommand\rows{{S}}
\newcommand\cols{{T}}

\newtheorem{theorem}{Theorem}
\newtheorem{corollary}{Corollary}
\newtheorem{proposition}{Proposition}

\begin{document}

\maketitle
{\def\thefootnote{}
\footnotetext{E-mail:
\texttt{mbelkin@ucsd.edu},
\texttt{djhsu@cs.columbia.edu},
\texttt{jixu@cs.columbia.edu}}}

\begin{abstract}
The ``double descent'' risk curve was proposed to qualitatively describe the out-of-sample prediction accuracy of variably-parameterized machine learning models. This article provides a precise mathematical analysis for the shape of this curve in two simple data models with the least squares/least norm predictor. Specifically, it is shown that the risk peaks when the number of features $p$ is close to the sample size $n$, but also that the risk decreases towards its minimum as $p$ increases beyond $n$. This behavior is contrasted with that of ``prescient'' models that select features in an \emph{a priori} optimal order.
\end{abstract}

\section{Introduction}

The ``double descent'' risk curve was proposed by \citet*{belkin2019reconciling} as a general way to qualitatively describe the out-of-sample prediction performance of variably-parameterized machine learning models.
This risk curve reconciles the classical bias-variance trade-off with the behavior of predictive models that interpolate training data, as observed for several model families (including neural networks) in a wide variety of applications (see Section~\ref{sec:related} for references).
In these studies, a predictive model with $p$ parameters is fit to a training sample of size $n$, and the test risk (i.e., out-of-sample error) is examined as a function of $p$.
When $p$ is below the sample size $n$ (for regression or binary classification), the test risk is governed by the usual bias-variance decomposition.
As $p$ is increased towards $n$, the training risk (i.e., in-sample error) is driven to zero, but the test risk shoots up, sometimes toward infinity. 
The classical bias-variance analysis identifies a ``sweet spot'' value of $p \in [0,n]$ at which the bias and variance are balanced to achieve low test risk.
However, in the ``modern regime'', as $p$ grows beyond $n$, the training risk remains zero, but the test risk decreases again, even when fitting noisy data, provided that the model is fit using a suitable inductive bias (e.g., least norm solution). In many (but not all) cases from~\citep{belkin2019reconciling}, the limiting risk as $p \to \infty$ is lower than what is achieved at the ``sweet spot'' value of $p$. 

In this article, we show that key aspects of the ``double descent'' risk curve can be observed with the least squares/least norm predictor in two simple random features models.
The first is a Gaussian model studied by \citet{breiman1983many} in the classical $p\leq n$ regime, while the second is a Fourier series model for functions on the circle.
In both cases, we prove that the risk is infinite around $p = n$, and decreases again as $p$ increases beyond $n$.
When the signal-to-noise ratio is high, the minimum risk is, in fact, achieved in the modern regime, when $p > n$.
Our results provide a precise mathematical analysis in a simple and tractable setting of the mechanism that was qualitatively described by~\citet{belkin2019reconciling}. 
In particular, it captures a key aspect of many practical over-parameterized models: that increasing the number of parameters to the maximum can lead to better performance.
We also establish some non-asymptotic concentration phenomena in the Gaussian model.

We note that in both of the models, the features are selected randomly, which makes them useful for studying scenarios where features are plentiful but individually too ``weak'' to be selected in an informed manner.
Such scenarios are commonplace in machine learning practice, and they should be contrasted with ``scientific'' scenarios where features are carefully designed or curated, as is often the case in scientific applications. 
For comparison, we give an example of  ``prescient'' feature selection, where the $p$  features \emph{a priori} known to be most useful are included in the model.
In this case, the optimal test risk is achieved at some $p \leq n$, which is consistent with the classical analysis of \citet{breiman1983many}.

\subsection{Related and concurrent works}
\label{sec:related}

The ``double descent'' risk curve was posited by \citet{belkin2019reconciling} to connect the classical bias-variance trade-off to behaviors observed in over-parameterized regimes for a variety of machine learning models.
The shape and features of the risk curve itself appear throughout in the literature in a number of contexts~\cite[e.g.,][]{vallet1989linear,opper1990ability,le1991eigenvalues,krogh1992generalization,bos1998dynamics,watkin1993statistical,advani2017high}; see also~\cite{loog2020brief} for a ``brief prehistory'' that focuses on the curious peak in the curve.
These prior works analyze the risk of linear classification and regression models and neural networks in high-dimensional asymptotic regimes.
Our analysis in the Gaussian model gives an exact expression for the risk for any finite sample size and number of parameters.

More recently, \citet{neal2018modern} observe that similar phenomena in neural networks can be explained by a variance reduction effect of increasing network width.
The transition from under- to over-parametrized regimes was recently analyzed by \citet*{spigler2018jamming} by drawing a connection to the physical phenomenon of ``jamming'' in a class of glassy systems.
Our analysis makes these ideas concrete and explicit in the context of simple regression models.
For instance, our analysis captures the transition from under- to over-parameterized regimes at a point where an inverse Wishart random matrix has no finite expectation.
It also allows us to compare the risks at any points in the curve and explain how the risk in the over-parameterized regime can be lower than any risk in the under-parameterized regime.

The initial version of this article~\citep{belkin2019two} appeared concurrently with the works of \citet{hastie2019surprises}, \citet{muthukumar2020harmless}, and
\citet{bartlett2020benign}, all of which also study the behavior of the least squares/least norm predictor in over-parameterized linear regression.
\citet{muthukumar2020harmless} focus on the well-specified scenario (essentially, $p = D$) and provide upper-bounds on the risk that go to zero as $p\to\infty$.
(A related variance analysis was carried out by \citet{neal2018modern}.)
\citet{hastie2019surprises} provide a much broader range of analyses in the high-dimensional asymptotic regime, including a ``misspecified'' setup that is related to ours.
Their analyses require weaker distributional assumptions than ours, owing to their reliance on asymptotic analysis.
(A special case of the results in the follow-up work by~\citet{xu2019number} further broadens the range of analyses to allow highly non-isotropic designs, but again only in the high-dimensional asymptotic regime.)
The analysis of \citeauthor{hastie2019surprises} also considers the effect of ridge regularization; in particular, they show that when the optimal level of regularization is used, the risk curve no longer shows the ``double descent'' shape.
Finally, \citet{bartlett2020benign} study non-asymptotic upper and lower bounds on the risk in the over-parameterized regime, and provide a characterization in terms of certain ``effective dimensions'' based on the tail of the eigenvalue sequence of the covariance operator.

\section{Gaussian model}
\label{sec:gaussian}

We consider a regression problem where the response $y$ is equal to a linear function $\vbeta = (\beta_1,\dotsc,\beta_D) \in \R^D$ of $D$ real-valued variables $\vx = (x_1,\dotsc,x_D)$ plus noise $\sigma\epsilon$:
\[
  y = \vx^\T \vbeta + \sigma\epsilon = \sum_{j=1}^D x_j \beta_j + \sigma\epsilon .
\]
Given $n$ iid copies $((\vx^{(i)},y^{(i)}))_{i=1}^n$ of $(\vx,y)$, we fit a linear model to the data only using a subset $\cols \subseteq [D] := \{1,\dotsc,D\}$ of $p := |\cols|$ variables.

Let $\vX := [ \vx^{(1)} | \dotsb | \vx^{(n)} ]^\T$ be the $n \times D$ design matrix, and let $\vy := (y^{(1)},\dotsc,y^{(n)})$ be the vector of responses.
For a subset $A \subseteq [D]$ and a $D$-dimensional vector $\vv$, we use $\vv_A := (v_j : j \in A)$ to denote its $|A|$-dimensional subvector of entries from $A$; we also use $\vX_A := [ \vx_A^{(1)} | \dotsb | \vx_A^{(n)} ]^\T$ to denote the $n \times |A|$ design matrix with variables from $A$.
For $A \subseteq [D]$, we denote its complement by $A^c := [D] \setminus A$.
Finally, $\|\cdot\|$ denotes the Euclidean norm.

We fit regression coefficients $\hat\vbeta = (\hat\beta_1,\dotsc,\hat\beta_D)$ with
\begin{equation*}
  \hat\vbeta_\cols := \vX_\cols^{\dag} \vy , \quad
  \hat\vbeta_{\cols^c} := \v0 .
\end{equation*}
Above, the symbol $^\dag$ denotes the Moore-Penrose pseudoinverse.
In other words, we use the solution to the normal equations $\vX_\cols^\T \vX_\cols \vv = \vX_\cols^\T \vy$ of least norm for $\hat\vbeta_\cols$ and force $\hat\vbeta_{\cols^c}$ to all-zeros.

In this section, our analysis assumes a model in which $(\vx,\epsilon)$ follows a standard multivariate Gaussian distribution.
This Gaussian model was also studied by \citet{breiman1983many}, although their analysis is restricted to the case where the number of variables used $p$ is always at most $n$; our analysis will also consider the $p \geq n$ regime.

\subsection{Prediction risk}
\label{sec:gaussian-risk}

We derive a formula for the (prediction) risk of $\hat\vbeta$ for an arbitrary choice of $p$ features $\cols \subseteq [D]$, and then examine this risk under particular selection models for $\cols$.

\begin{theorem}
  \label{thm:risk-noise}
  Assume the distribution of $\vx$ is the standard normal in $\R^D$, $\epsilon$ is a standard normal random variable independent of $\vx$, and $y = \vx^\T\vbeta + \sigma\epsilon$ for some $\vbeta \in \R^D$ and $\sigma>0$.
  Pick any $p \in \{0,\dotsc,D\}$ and $\cols \subseteq [D]$ of cardinality $p$.
  The risk of $\hat\vbeta$, where $\hat\vbeta_\cols = \vX_\cols^\dag \vy$ and $\hat\vbeta_{\cols^c} = \v0$, is
  \begin{equation*}
    \E[ (y - \vx^\T\hat\vbeta)^2 ]
    =
    \begin{cases}
      (\|\vbeta_{\cols^c}\|^2 + \sigma^2) \cdot \left( 1 + \frac{p}{n-p-1} \right) & \text{if $p \leq n-2$} ; \\
      +\infty & \text{if $n-1 \leq p \leq n+1$} ; \\
      \|\vbeta_\cols\|^2 \cdot \left( 1 - \frac{n}{p} \right) + (\|\vbeta_{\cols^c}\|^2 + \sigma^2) \cdot \left( 1 + \frac{n}{p-n-1} \right) & \text{if $p \geq n+2$} .
    \end{cases}
  \end{equation*}
\end{theorem}
The proof of \Cref{thm:risk-noise} is not hard, we give the details in \Cref{sec:proof}.
We now turn to the risk of $\hat\vbeta$ under a random selection model for $\cols$.
\begin{corollary}
  \label{cor:risk-noise-random}
  Let $\cols$ be a uniformly random subset of $[D]$ of cardinality $p$.
  In the setting of \Cref{thm:risk-noise}, the risk of $\hat\vbeta$ (taking expectation with respect to the random choice of $\cols$ in addition to the random design matrix and response vector) satisfies
  \begin{equation*}
    \E[ (y - \vx^\T\hat\vbeta)^2 ]
    =
    \begin{cases}
      \left( \left( 1 - \frac{p}{D} \right) \cdot \|\vbeta\|^2 + \sigma^2 \right) \cdot \left( 1 + \frac{p}{n-p-1} \right) & \text{if $p \leq n-2$} ; \\
      \|\vbeta\|^2 \cdot \left( 1 - \frac{n}{D} \cdot \left( 2 - \frac{D-n-1}{p-n-1} \right) \right) + \sigma^2 \cdot \left( 1 + \frac{n}{p-n-1} \right) & \text{if $p \geq n+2$} .
    \end{cases}
  \end{equation*}
\end{corollary}
\begin{proof}
  Since $\cols$ is a uniformly random subset of $[D]$ of cardinality $p$,
  \begin{equation*}
    \E[ \|\vbeta_\cols\|^2 ] = \frac{p}{D} \cdot \|\vbeta\|^2 , \quad
    \E[ \|\vbeta_{\cols^c}\|^2 ] = \left( 1 - \frac{p}{D} \right) \cdot \|\vbeta\|^2 .
  \end{equation*}
  Plugging into \Cref{thm:risk-noise} completes the proof.
\end{proof}

\begin{figure}
  \centering
  \includegraphics[width=0.5\textwidth]{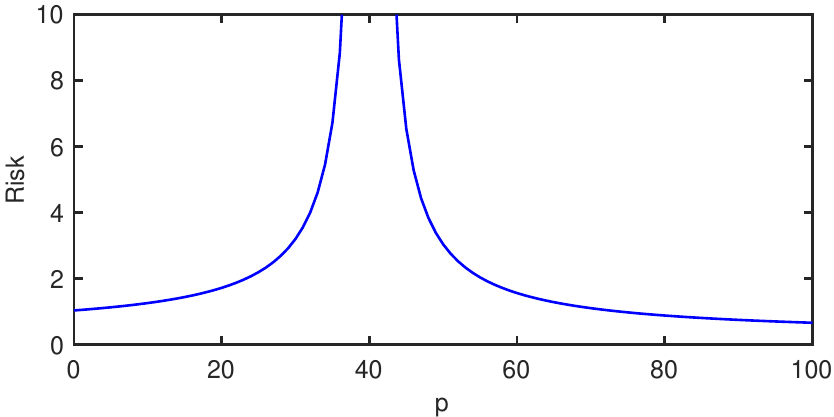}
  \caption{Plot of risk $\E[ (y - \vx^\T\hat\vbeta)^2 ]$ as a function of $p$, under the random selection model of $\cols$. Here, $\|\vbeta\|^2 = 1$, $\sigma^2 = 1/25$, $D = 100$, and $n = 40$.}
  \label{fig:risk-plot}
\end{figure}

Thus, assuming $D>n+1$, we observe that the risk first \emph{increases} with $p$ up to the ``interpolation threshold'' ($p=n$), after which the risk \emph{decreases} with $p$.
Moreover, when the signal-to-noise ratio $\|\vbeta\|^2/\sigma^2$ is larger than $D/(D-n-1)$, the risk is smallest at $p=D$; in particular, it is smaller than the risk at any $p \leq n$.
This is the ``double descent'' risk curve where the first ``descent'' is degenerate (i.e., the ``sweet spot'' that balances bias and variance is at $p=0$).
See \Cref{fig:risk-plot} for an illustration.

\begin{figure}
  \centering
  \includegraphics[width=0.5\textwidth]{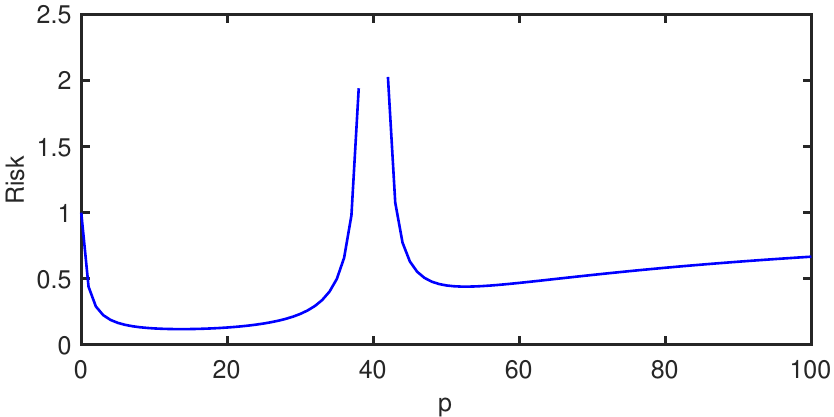}
  \caption{Plot of risk $\E[ (y - \vx^\T\hat\vbeta)^2 ]$ as a function of $p$, under the ``prescient'' selection model of $\cols$.
  Here, $\|\vbeta\|^2 = 1$, $\beta_j^2 \propto 1/j^2$, $\sigma^2 = 1/25$, $D = 100$, and $n = 40$.}
  \label{fig:risk-plot2}
\end{figure}

It is worth pointing out that the behavior under the random selection model of $\cols$ can be very different from that under a deterministic model of $\cols$.
Consider including variables in $\cols$ by decreasing order of $\beta_j^2$---a kind of ``prescient'' selection model studied by \citet{breiman1983many}.
The behavior of the risk as a function of $p$, illustrated in \Cref{fig:risk-plot2}, reveals a striking difference between the random selection model and the ``prescient'' selection model.

\subsection{Proof of \Cref{thm:risk-noise}}
\label{sec:proof}

Recall that $\vx$ is assumed to follow a standard normal distribution in $\R^D$.
Since $\vx$ is isotropic (i.e., zero mean and identity covariance), the mean squared prediction error of any $\vbeta' \in \R^D$ can be written as
\begin{equation*}
  \E[ (y - \vx^\T \hat{\vbeta})^2 ]
  = \sigma^2 + \|\vbeta - \hat{\vbeta}\|^2
  = \sigma^2 + \|\vbeta_{\cols^c} - \hat{\vbeta}_{\cols^c}\|^2
  + \|\vbeta_\cols - \hat{\vbeta}_\cols\|^2
  .
\end{equation*}
Since $\hat\vbeta_{\cols^c} = \v0$, it follows that the risk of $\hat\vbeta$ is
\begin{equation*}
  \E[ (y - \vx^\T \hat\vbeta)^2 ] = \sigma^2 + \|\vbeta_{\cols^c}\|^2 + \E[ \|\vbeta_\cols - \hat\vbeta_\cols\|^2 ] .
\end{equation*}

\paragraph{Classical regime.}
The risk of $\hat\vbeta$ was computed by \citet{breiman1983many} in the regime where $p \leq n$:
\begin{equation*}
  \E[ (y - \vx^\T \hat\vbeta)^2 ]
  =
  \begin{cases}
    (\|\vbeta_{\cols^c}\|^2 + \sigma^2) \cdot \left( 1 + \frac{p}{n-p-1} \right)
    & \text{if $p \leq n-2$} ; \\
    +\infty
    & \text{if $p \in \{ n-1, n \}$} .
  \end{cases}
\end{equation*}

\paragraph{Interpolating regime.}
We consider the regime where $p \geq n$.
Recall that the pseudoinverse of $\vX_\cols$ can be written as $\vX_\cols^{\dag} = \vX_\cols^\T (\vX_\cols\vX_\cols^\T)^\dag$.
Thus, letting $\veta := \vy - \vX_\cols\vbeta_\cols$,
\begin{align*}
  \vbeta_\cols - \hat\vbeta_\cols
  & = \vbeta_\cols - \vX_\cols^\T (\vX_\cols\vX_\cols^\T)^\dag \vy \\
  & = \vbeta_\cols - \vX_\cols^\T (\vX_\cols\vX_\cols^\T)^\dag (\vX_\cols \vbeta_\cols + \veta) \\
  & = (\vI - \vX_\cols^\T (\vX_\cols\vX_\cols^\T)^\dag \vX_\cols) \vbeta_\cols - \vX_\cols^\T (\vX_\cols\vX_\cols^\T)^\dag \veta .
\end{align*}
On the right hand side, the first term $(\vI - \vX_\cols^\T (\vX_\cols\vX_\cols^\T)^\dag \vX_\cols) \vbeta_\cols$ is the orthogonal projection of $\vbeta_\cols$ onto the null space of $\vX_\cols$, while the second term $-\vX_\cols^\T (\vX_\cols\vX_\cols^\T)^\dag \veta$ is a vector in the row space of $\vX_\cols$.
By the Pythagorean theorem, the squared norm of their sum is equal to the sum of their squared norms, so
\begin{equation*}
  \|\vbeta_\cols - \hat\vbeta_\cols\|^2
  = \|(\vI - \vX_\cols^\T (\vX_\cols\vX_\cols^\T)^\dag \vX_\cols) \vbeta_\cols\|^2
  + \|\vX_\cols^\T (\vX_\cols\vX_\cols^\T)^\dag \veta\|^2 .
\end{equation*}
We analyze the expected values of these two terms by exploiting properties of the standard normal distribution.

\begin{description}
  \item[First term.]
    Note that $\vPi_\cols := \vX_\cols^\T (\vX_\cols\vX_\cols^\T)^\dag \vX_\cols$ is the orthogonal projection matrix for the row space of $\vX_\cols$.
    So, by the Pythagorean theorem, we have
    \begin{equation*}
      \|(\vI - \vX_\cols^\T (\vX_\cols\vX_\cols^\T)^\dag \vX_\cols) \vbeta_\cols\|^2
      = \|\vbeta_\cols\|^2 - \|\vPi_\cols\vbeta_\cols\|^2 .
    \end{equation*}
    By rotational symmetry of the standard normal distribution, it follows that
    \begin{equation*}
      \E[ \|\vPi_\cols\vbeta_\cols\|^2 ]
      = \|\vbeta_\cols\|^2 \cdot \frac{n}{p} .
    \end{equation*}
    Therefore
    \begin{equation*}
      \E[\|(\vI - \vX_\cols^\T (\vX_\cols\vX_\cols^\T)^\dag \vX_\cols) \vbeta_\cols\|^2]
      = \|\vbeta_\cols\|^2 \cdot \left( 1 - \frac{n}{p} \right) .
    \end{equation*}

  \item[Second term.]
    We use the ``trace trick'' to write
    \begin{equation*}
      \|\vX_\cols^\T (\vX_\cols\vX_\cols^\T)^\dag \veta\|^2
      = \tr( (\vX_\cols\vX_\cols^\T)^\dag (\vX_\cols\vX_\cols^\T) (\vX_\cols\vX_\cols^\T)^\dag \veta \veta^\T )
      = \tr( (\vX_\cols\vX_\cols^\T)^\dag \veta \veta^\T )
    \end{equation*}
    where the second equality holds almost surely because $\vX_\cols\vX_\cols^\T$ is almost surely invertible.
    Since $\vx_\cols^\T\vbeta_\cols$ and $\vx_{\cols^c}^\T\vbeta_{\cols^c} + \sigma\epsilon$ are uncorrelated, it follows that
    \begin{equation*}
      \E[ \|\vX_\cols^\T (\vX_\cols\vX_\cols^\T)^\dag \veta\|^2 ]
      = \tr( \E[ (\vX_\cols\vX_\cols^\T)^\dag ] \E[ \veta \veta^\T ] ) .
    \end{equation*}
    The distribution of $\veta$ is normal with mean zero and covariance $(\|\vbeta_{\cols^c}\|^2 + \sigma^2) \cdot \vI \in \R^{n \times n}$, so
    \begin{equation*}
      \E[ \veta \veta^\T ]
      = (\|\vbeta_{\cols^c}\|^2 + \sigma^2) \cdot \vI .
    \end{equation*}
    The distribution of $\vP := (\vX_\cols\vX_\cols^\T)^\dag$ is inverse-Wishart with identity scale matrix $\vI \in \R^{n \times n}$ and $p$ degrees-of-freedom.
    Each diagonal entry $P_{i,i}$ of $\vP$, for $i=1,\dotsc,n$, has a reciprocal that follows the $\chi^2$ distribution with $p - n + 1$ degrees-of-freedom.
    Hence $\E[ P_{i,i} ] = 1/(p-n-1)$ if $p \geq n+2$ and $\E[ P_{i,i} ] = +\infty$ if $p \in \{n,n+1\}$.
    Therefore
    \begin{equation*}
      \tr( \E[ (\vX_\cols\vX_\cols^\T)^\dag ] )
      =
      \begin{cases}
        \frac{n}{p-n-1} & \text{if $p \geq n+2$} ; \\
        +\infty & \text{if $p \in \{n, n+1\}$} .
      \end{cases}
    \end{equation*}
    We conclude that
    \begin{equation*}
      \E[ \|\vX_\cols^\T (\vX_\cols\vX_\cols^\T)^\dag \veta\|^2 ]
      =
      \begin{cases}
        (\|\vbeta_{\cols^c}\|^2 + \sigma^2) \cdot \frac{n}{p-n-1} & \text{if $p \geq n+2$} ; \\
        +\infty & \text{if $p \in \{n, n+1\}$} .
      \end{cases}
    \end{equation*}

\end{description}
Combining the first and second terms gives the claimed expression for the risk.
\qed

\subsection{Concentration}
\label{sec:conc}

We briefly consider the measure concentration of $\|\vbeta - \hat\vbeta\|^2$.
\begin{theorem}
  \label{thm:conc}
  Consider the setting from \Cref{thm:risk-noise}, and fix any $\epsilon \in (0,1)$.
  If $\alpha := p/n < 1$, then
  \begin{align*}
    \|\vbeta - \hat\vbeta\|^2 & \in
    (\|\vbeta_{T^c}\|^2 + \sigma^2) \left( 1 +  \left(\frac{1\pm\epsilon}{1\mp\epsilon}\right) \frac{p}{n-p+1} \right)
  \end{align*}
  with probability at least
  \begin{equation*}
    1 - 2\exp\left( -\frac{p\epsilon^4(\sqrt{\alpha^{-1}}-1)^2}{24((2-\epsilon)\sqrt{\alpha^{-1}}+\epsilon)^2} \right)
    - 2\exp\left( -\frac{p(1-\epsilon)^2(\sqrt{\alpha^{-1}}-1)^2}{2} \right)
    - 2p\exp\left( -\frac{p(\alpha^{-1}-1)\epsilon^2}{24} \right) .
  \end{equation*}
  If $\alpha > 1$, then
  \begin{align*}
    \|\vbeta - \hat\vbeta\|^2 & \in \|\vbeta_T\|^2 \left( 1 - (1\pm\epsilon)\frac{n}{p} \right)
    + (\|\vbeta_{T^c}\|^2 + \sigma^2) \left( 1 +  \left(\frac{1\pm\epsilon}{1\mp\epsilon}\right) \frac{n}{p-n+1} \right)
  \end{align*}
  with probability at least
  \begin{equation*}
    1 - 2\exp\left( -\frac{n\epsilon^2}{12} \right)
    - 2\exp\left( -\frac{n\epsilon^4(\sqrt{\alpha}-1)^2}{24((2-\epsilon)\sqrt{\alpha}+\epsilon)^2} \right)
    - 2\exp\left( -\frac{n(1-\epsilon)^2(\sqrt{\alpha}-1)^2}{2} \right)
    - 2n\exp\left( -\frac{n(\alpha-1)\epsilon^2}{24} \right) .
  \end{equation*}
\end{theorem}
The proof is given in \Cref{app:conc}.
The main idea for the $p>n$ case is as follows.
From the proof of \Cref{thm:risk-noise}, we have the decomposition
\begin{align*}
  \|\vbeta_T - \hat\vbeta_T\|^2
  & = \|(\vI - \vPi_T) \vbeta_T\|^2 + \|\vX_T^\T(\vX_T\vX_T^\T)^\dag\veta\|^2 .
\end{align*}
The first term $\|(\vI - \vPi_T) \vbeta_T\|^2$ is the squared distance from $\vbeta_T$ to a uniformly random $n$-dimensional subspace of $\R^p$.
This squared distance has the same distribution as the squared distance from a uniformly random vector of length $\|\vbeta_T\|$ to a fixed $n$-dimensional subspace of $\R^p$.
Thus measure concentration on the unit sphere can be used here.
The second term $\|\vX_T^\T(\vX_T\vX_T^\T)^\dag\veta\|^2$ is a (random) quadratic form in the Gaussian random vector $\veta$.
Gaussian concentration is readily applied after controlling the spectral properties of the Wishart random matrix $\vX_T\vX_T^\T$.
(The $p<n$ case is similar to the analysis of this second term.)

The same arguments can be used to give fixed-level confidence bounds; see \Cref{prop:confidence} in \Cref{app:conf}.

Finally, it is also possible to compare $\|\vbeta_T\|^2$ to $(p/D)\|\vbeta\|^2$ (and $\|\vbeta_{T^c}\|^2$ to $(1-p/D)\|\vbeta\|^2$) under the random selection model of $T$ from \Cref{cor:risk-noise-random} using concentration inequalities for sampling without replacement \citep[see, e.g.,][for a discussion]{bardenet2015concentration}.
The following is a simple consequence of Proposition~1.4 of \citep{bardenet2015concentration}.
\begin{proposition}
  \label{prop:sampling}
  For any $t>0$, with probability at least $1-2e^{-t}$,
  \begin{align*}
    \left| \|\vbeta_T\|^2 - \frac{p}{D} \|\vbeta\|^2 \right|
    = \left| \|\vbeta_{T^c}\|^2 - \left(1 - \frac{p}{D}\right) \|\vbeta\|^2 \right|
    & \leq \|\vbeta\|^2 \left( \sqrt{2\left(\mu^2-\frac1D\right) \min\left\{ \frac{p}{D} , 1 - \frac{p}{D} \right\} t} + \frac{2\mu^2 t}{3} \right) .
  \end{align*}
  where $\mu := \max_{i \in [D]} |\beta_i| / \|\vbeta\|$.
\end{proposition}
The proof is in \Cref{app:sampling}.
The crucial parameter $\mu$ has range $[1/\sqrt{D},1]$.
It is small when there are many relevant ``weak'' features, each with a relatively small coefficient in $\vbeta$; conversely, it is large when $\vbeta$ is concentrated on a sparse subset of features.

\section{Fourier series model}
\label{sec:fourier}

In this section, we consider a noise-free Fourier series model, which can be regarded as a one-dimensional version of the random Fourier features model studied by~\citet{rahimi2008random} for functions defined on the unit circle.

Let $\vF \in \C^{D \times D}$ denote the $D \times D$ discrete Fourier transform matrix: its $(i,j)$-th entry is
\begin{equation*}
  F_{i,j} = \frac1{\sqrt{D}} \omega^{(i-1)(j-1)} ,
\end{equation*}
where $\omega := \exp(-2\pi\mathrm{i}/D)$ is a primitive root of unity.
Let $\vmu := \vF\vbeta$ for some $\vbeta \in \C^D$.
Consider the following observation model:
\begin{enumerate}
  \item $\rows$ and $\cols$ are independent random subsets of $[D]$.
    For any $i \in [D]$, the membership of $i$ in $\rows$ (respectively, $\cols$) is determined by an independent Bernoulli variable with mean $\rho_n:=n/D$ (respectively, $\rho_p:=p/D$).

  \item We observe the $n \times p$ design matrix $\vF_{\rows,\cols}$ and $n$-dimensional vector of responses $\vmu_\rows$.
    Here, $\vF_{\rows,\cols}$ is the submatrix of $\vF$ with rows from $\rows$ and columns from $\cols$, and $\vmu_\rows$ is the subvector of $\vmu$ of entries from $\rows$.

\end{enumerate}
We fit regression coefficients $\hat\vbeta = (\hat\beta_1,\dotsc,\hat\beta_D)$ with
\begin{equation*}
  \hat\vbeta_\rows := \vF_{\rows,\cols}^{\dag} \vmu_\rows , \quad
  \hat\vbeta_{\rows^c} := \v0 .
\end{equation*}

One important property of the discrete Fourier transform matrix that we use is that the matrix $\vF_{A,B}$ has rank $\min\{|A|,|B|\}$ for any $A, B \subseteq [D]$.
This is a consequence of the fact that $\vF$ is Vandermonde.
Thus, we have 
\begin{equation*}
  \vF_{\rows,\cols}^\dag = \left\{\begin{aligned}&\vF_{\rows,\cols}^* (\vF_{\rows,\cols}\vF_{\rows,\cols}^*)^{-1}, &&|T|\geq |S|\\
  &(\vF_{\rows,\cols}^*\vF_{\rows,\cols})^{-1}\vF_{\rows,\cols}^*,&&|T|\leq |S|\\
  \end{aligned}\right..
\end{equation*}

In the remainder of this section, we analyze the risk of $\hat\vbeta$ under a random model for $\vbeta$, where
\begin{equation*}
  \E[ \vbeta \vbeta^* ] = \frac{1}{D} \cdot \vI
\end{equation*}
(which implies $\E[ \|\vbeta\|^2 ] = 1$).
The random choice of $\vbeta$ is independent of $\rows$ and $\cols$.
Considering the risk under this random model for $\vbeta$ is a form of average-case analysis.
For simplicity, we only consider the regime where $\rho_p > \rho_n$.

\begin{figure}
  \centering
  \includegraphics[width=0.5\textwidth]{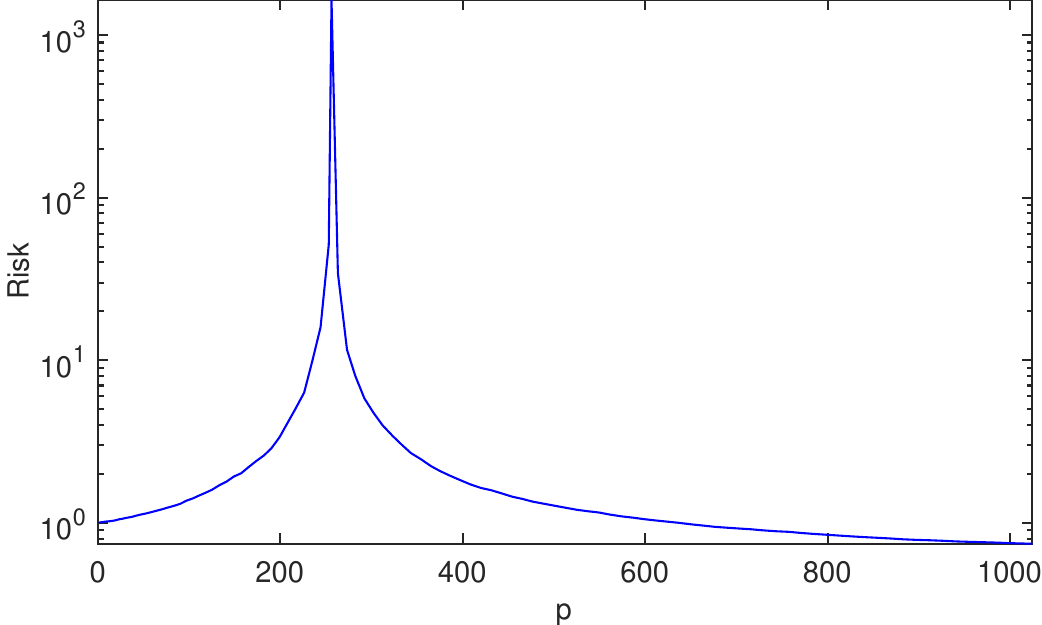}
  \caption{Plot of risk as a function of $p$ in the Fourier series model. Here, $\vbeta$ was chosen uniformly at random (once) from the unit sphere in $\R^D$ for $D=1024$. We then computed $\hat\vbeta$ from $10$ independent random choices of $\rows$ (with $n = 256$) and $\cols$ and plotted the average value of $\|\vbeta-\hat\vbeta\|^2$.}
  \label{fig:fourier-plot}
\end{figure}

Following the arguments from \Cref{sec:gaussian-risk}, we have
\begin{align*}
  \|\vbeta - \hat\vbeta\|^2
  & = \|\vbeta_{\rows^c}\|^2
  + \|(\vI - \vF_{\rows,\cols}^\dag \vF_{\rows,\cols}) \vbeta_\rows\|^2
  + \|\vF_{\rows,\cols}^\dag \vF_{\rows,\cols^c} \vbeta_{\rows^c}\|^2 \\
  & = \|\vbeta\|^2
  - \|\vF_{\rows,\cols}^\dag \vF_{\rows,\cols} \vbeta_\rows\|^2
  + \|\vF_{\rows,\cols}^\dag \vF_{\rows,\cols^c} \vbeta_{\rows^c}\|^2 .
\end{align*}
Now we take (conditional) expectations with respect to $\vbeta$, given $\rows$ and $\cols$:
\begin{equation}
  \E[ \|\vbeta - \hat\vbeta\|^2 \mid \rows,\cols ]
  = 1
  - \frac{1}{D} \cdot \tr((\vF_{\rows,\cols}^\dag\vF_{\rows,\cols})^* (\vF_{\rows,\cols}^\dag\vF_{\rows,\cols}))
  + \frac{1}{D} \cdot \tr((\vF_{\rows,\cols}^\dag\vF_{\rows,\cols^c})^* (\vF_{\rows,\cols}^\dag\vF_{\rows,\cols^c})) .
  \label{eq:fourier-risk-decomp}
\end{equation}
Since $\vF_{\rows,\cols}$ has rank $\min\{|\rows|,|\cols|\}$, the first trace expression is equal to
\begin{equation*}
  \tr( (\vF_{\rows,\cols}^\dag \vF_{\rows,\cols})^*(\vF_{\rows,\cols}^\dag \vF_{\rows,\cols}) ) = \min\{ |\rows|, |\cols| \} .
\end{equation*}
For the second trace expression, we use the explicit formula for $\vF_{\rows,\cols}^\dag$ and the fact that $\vF_{\rows,\cols} \vF_{\rows,\cols}^* + \vF_{\rows,\cols^c} \vF_{\rows,\cols^c}^* = \vI$ to obtain
\begin{align*}
  \tr( (\vF_{\rows,\cols}^\dag \vF_{\rows,\cols^c})^*(\vF_{\rows,\cols}^\dag \vF_{\rows,\cols^c}) )
  & = \tr( \vF_{\rows,\cols^c}^* (\vF_{\rows,\cols} \vF_{\rows,\cols}^*)^{-1} \vF_{\rows,\cols^c} ) \\
  & = \tr( \vF_{\rows,\cols^c}^* (\vI - \vF_{\rows,\cols^c} \vF_{\rows,\cols^c}^*)^{-1} \vF_{\rows,\cols^c} ) \\
  & = \tr( (\vI - \vF_{\rows,\cols^c} \vF_{\rows,\cols^c}^*)^{-1} \vF_{\rows,\cols^c} \vF_{\rows,\cols^c}^* ) \\
  & = \sum_{i=1}^{\min\{|\rows|,|\cols|\}} \frac{\lambda_i}{1 - \lambda_i} \\
  & = -\min\{|\rows|,|\cols|\} + \sum_{i=1}^{\min\{|\rows|,|\cols|\}} \frac{1}{1 - \lambda_i} ,
\end{align*}
where the $\lambda_i \in [0,1]$ are the eigenvalues of $\vF_{\rows,\cols^c} \vF_{\rows,\cols^c}^*$.
Therefore, from \Cref{eq:fourier-risk-decomp}, we have
\begin{equation*}
  \E[ \|\vbeta - \hat\vbeta\|^2 ]
  = 1 - 2\E\min\left\{\frac{|\rows|}{D},\frac{|\cols|}{D}\right\} + \frac{n}{D} \cdot \underbrace{\E\left[ \frac{1}{n} \sum_{i=1}^{\min\{|\rows|,|\cols|\}} \frac1{1-\lambda_i} \right]}_{(*)} .
\end{equation*}
To determine the asymptotic behavior of $(*)$, we use a recent result of \citet{farrell2011limiting}:
\begin{equation*}
  (*) \to \frac{\rho_p \cdot (1-\rho_n)}{\rho_p - \rho_n}
\end{equation*}
as $D,n,p \to \infty$ with $\rho_n = n/D$ and $\rho_p = p/D$ held fixed.
Further, under this limit, we have
\[
  \E \min\left\{\frac{|\rows|}{D}, \frac{|\cols|}{D} \right\} \rightarrow \rho_n
\]
since $\rho_p \geq \rho_n$.
Hence we have the following:
\begin{theorem}
  \label{thm:fourier}
  Assume the setting as above, with  $D,n,p \to \infty$ and $\rho_n = n/D$ and $\rho_p = p/D$ held fixed. Then 
\begin{equation*}
  \lim\,\E\,\left[ \|\vbeta - \hat\vbeta\|^2 \right]
  = 1 -\frac{n}{D}\left(2 - \frac{p(1-n/D)}{p-n}\right). 
\end{equation*}
\end{theorem}
Note that the right-hand side in the equation from \Cref{thm:fourier} is well-defined in the limit because the ratios $\rho_n,\rho_p$ are fixed.
It diverges to $+\infty$ when $\rho_p$ is close to $\rho_n$, and decreases as $\rho_p$ approaches $1$. This is the same behavior as in the Gaussian model from \Cref{sec:gaussian} with random feature selection; we depict a non-asymptotic instantiation of it in \Cref{fig:fourier-plot}.

\section{Discussion}

Our analysis shows that when features are chosen in an uninformed manner, it may be optimal to choose as many as possible---even more than the number of data---rather than limit the number to that which balances bias and variance as suggested by classical analyses. This choice is simple, both conceptually and algorithmically (although it may incur a computational penalty for processing large numbers of parameters), and avoids the need for precise control of regularization parameters. 
It is reflective of the practice in modern machine learning applications like image and speech recognition, where signal processing-based features are individually weak but in great abundance, and models that use all of the features, notably neural networks,  are highly successful.
This stands in contrast to the ``scientific'' scenarios with informed selection of features; for example, in many science and medical applications, features are purposefully chosen based on the detailed understanding of the underlying phenomena. 
As illustrated by the ``prescient''  model that selects  the best features, in that case choosing the number of features to balance bias and variance can be better than incurring the costs that come with using all of the features.

Finally we remark, that there appears to be a sharp divide between the classical  analyses of statistics and machine learning in  $p<n$ regimes and the modern ``weak but plentiful features'' interpolating settings. While the former are deeply explored, an understanding of the latter is only starting to emerge. 
It is clear that the best practices for model and feature selection  depend crucially on the  regime of the application.

\subsubsection*{Acknowledgements}
We thank the anonymous referees for their remarks and suggestions (which, in particular, led to the inclusion of \Cref{sec:conc}).
This work was carried out in part while MB was at The Ohio State University.
This research was supported by NSF CCF-1740833 and IIS-1815697 awards, a Sloan Research Fellowship, a Google Faculty Award, and a Cheung-Kong Graduate School of Business Fellowship.

\printbibliography

\appendix

\section{Proof of \Cref{thm:conc}}
\label{app:conc}

We first consider $p>n$ (i.e., $\alpha>1$).
From the proof of \Cref{thm:risk-noise}, we have the decomposition
\begin{align*}
  \|\vbeta_T - \hat\vbeta_T\|^2
  & = \|(\vI - \vPi_T) \vbeta_T\|^2 + \|\vX_T^\T(\vX_T\vX_T^\T)^\dag\veta\|^2 ,
\end{align*}
where $\vPi_T$ is the orthogonal projection matrix for the row space of $\vX_T$, and $\veta$ is normal with mean zero and covariance $(\|\vbeta_{T^c}\|^2 + \sigma^2) \vI$ and independent of $\vX_T$.
By symmetry of the standard normal distribution, the first term $\|(\vI - \vPi_T) \vbeta_T\|^2$ is the squared distance from $\vbeta_T$ to a uniformly random $n$-dimensional subspace of $\R^p$.
This squared distance has the same distribution as the squared distance from a uniformly random vector of length $\|\vbeta_T\|$ to a fixed $n$-dimensional subspace of $\R^p$.
This argument was also used by \citet{dasgupta2003elementary} in their proof of the Johnson-Lindenstrauss lemma.
By Lemma~2.2 from \citep{dasgupta2003elementary}, we have for any $\epsilon \in (0,1)$,
\begin{align*}
  \Pr\left[ \|(\vI - \vPi_T) \vbeta_T\|^2 \not\in \left( 1 - (1\pm\epsilon)\frac{n}{p} \right) \|\vbeta_T\|^2 \right] & \leq 2\exp\left( -\frac{n\epsilon^2}{12} \right) .
\end{align*}
The second term $\|\vX_T^*(\vX_T\vX_T^*)^\dag\veta\|^2$ is a (random) quadratic form in $\veta$.
Let $\vK_T := \vX_T\vX_T^*$, which is non-singular almost surely.
By Lemma~4 from \citep{dasgupta2000learning}, we have for any $\epsilon \in (0,1)$,
\begin{align*}
  \Pr\left[ \|\vX_T^*(\vX_T\vX_T^*)^\dag\veta\|^2 \not\in (1\pm\epsilon)(\|\vbeta_{T^c}\|^2 + \sigma^2) \tr(\vK_T^{-1}) \mid \text{$\vK_T$ non-singular} \right]
  & \leq 2\exp\left( -\frac{n\epsilon^2}{24\kappa(\vX_T)^2} \right) ,
\end{align*}
where $\kappa(\vX_T) = \sigma_{\max}(\vX_T) / \sigma_{\min}(\vX_T)$ is the ratio of the largest singular value of $\vX_T$ to the smallest singular value of $\vX_T$.
For any $t>0$,
\begin{align*}
  \Pr\left[ \sigma_{\max}(\vX_T) \geq \sqrt{p} + (1+t)\sqrt{n} \right] & \leq \exp(-nt^2/2) , \\
  \Pr\left[ \sigma_{\min}(\vX_T) \leq \sqrt{p} - (1+t)\sqrt{n} \right] & \leq \exp(-nt^2/2) .
\end{align*}
These inequalities follow from Gaussian comparison inequalities and concentration of measure on the sphere and in Gaussian space~\citep[see, e.g.,][]{rudelson2009smallest,vershynin2018high}.
Therefore, for $p > (1+t)^2 n$,
\begin{align*}
  \Pr\left[ \kappa(\vX_T)^2 \geq \left( \frac{\sqrt{p} + (1+t)\sqrt{n}}{\sqrt{p} - (1+t)\sqrt{n}} \right)^2 \right]
  & \leq 2\exp\left( - \frac{nt^2}{2} \right) .
\end{align*}
Finally, observe that $1/(\vK_T^{-1})_{i,i}$ has a $\chi^2$-distribution with $p-n+1$ degrees of freedom.
Therefore, again using Lemma~4 from \citep{dasgupta2000learning} and a union bound, we have for any $\epsilon \in (0,1)$,
\begin{align*}
  \Pr\left[ \tr(\vK_T^{-1}) \notin \frac{n}{p-n+1} \cdot \frac1{1\mp\epsilon} \right ]
  & \leq 2n \exp\left( -\frac{(p-n+1)\epsilon^2}{24} \right) .
\end{align*}
Putting these probability inequalities together (with $t=(1-\epsilon)(\sqrt{\alpha}-1)$) completes the proof for $p>n$.

Now we consider $p<n$ (i.e., $\alpha<1$).
We have
\begin{align*}
  \hat\vbeta_T & = (\vX_T^\T\vX_T)^\dag \vX_T^\T(\vX_T\vbeta_T + \veta) .
\end{align*}
The matrix $\vX_T^\T\vX_T$ is non-singular almost surely, so $\|\hat\vbeta_T - \vbeta\|^2 = \veta^\T (\vX_T \vX_T^\T)^\dag \veta = \veta^\T \vK_T^\dag \veta$ also holds almost surely.
Note that $\vK_T$ has the same eigenvalues as $\vX_T^\T\vX_T$, and hence $\vK_T^\dag$ has the same eigenvalues as $(\vX_T^\T\vX_T)^{-1}$.
Therefore, following essentially the same arguments as above for handling $\|\vX_T^\T(\vX_T\vX_T^\T)^\dag\veta\|^2$ (but switching the roles of $p$ and $n$, and hence replacing $\alpha$ with $\alpha^{-1}$) completes the proof for $p<n$.
\qed

\section{Confidence bounds}
\label{app:conf}

Fixed-level confidence bounds can be immediately derived from the probability inequalities in \Cref{app:conc}.
\begin{proposition}
  \label{prop:confidence}
  Consider the setting from \Cref{thm:risk-noise} and fix any $\delta \in (0,1)$.
  If $p < n$, then with probability at least $1-\delta$,
  \begin{multline*}
    \|\vbeta_T - \hat\vbeta_T\|^2
    \in \left( 1\pm \frac{1 + \sqrt{\frac{p}{n}} + \sqrt{\frac{2\ln(8/\delta)}{n}}}{1 - \sqrt{\frac{p}{n}} - \sqrt{\frac{2\ln(8/\delta)}{n}}} \cdot \sqrt{\frac{48\ln(256/\delta)}{p}} \right) (\|\vbeta_{T^c}\|^2 + \sigma^2) \cdot \frac{p}{n-p+1} \cdot \frac1{1 \mp \sqrt{\frac{24\ln(8p/\delta)}{n-p+1}}} .
  \end{multline*}
  If $p > n$, then with probability at least $1-\delta$,
  \begin{multline*}
    \|\vbeta_T - \hat\vbeta_T\|^2
    \in \left( 1 - \left( 1 \pm \sqrt{\frac{12\ln(8/\delta)}{n}} \right) \frac{n}{p} \right) \|\vbeta_T\|^2 \\
    + \left( 1\pm \frac{1 + \sqrt{\frac{n}{p}} + \sqrt{\frac{2\ln(8/\delta)}{p}}}{1 - \sqrt{\frac{n}{p}} - \sqrt{\frac{2\ln(8/\delta)}{p}}} \cdot \sqrt{\frac{48\ln(256/\delta)}{n}} \right) (\|\vbeta_{T^c}\|^2 + \sigma^2) \cdot \frac{n}{p-n+1} \cdot \frac1{1 \mp \sqrt{\frac{24\ln(8n/\delta)}{p-n+1}}} .
  \end{multline*}
  In the expressions above, we assume $n$ and $p$ are large enough (perhaps in relation to each other) so that all denominators are positive.
\end{proposition}

\section{Proof of \Cref{prop:sampling}}
\label{app:sampling}

Let $X_1,\dotsc,X_p$ denote a random sample of cardinality $p$ from the finite population $(\beta_1^2, \dotsc, \beta_D^2)$, drawn without replacement, so that $\|\vbeta_T\|^2 = \sum_{j=1}^p X_j$.
Since $\|\vbeta_{T^c}\|^2 = \|\vbeta\|^2 - \|\vbeta_T\|^2$, we have
\begin{align*}
  \left| \|\vbeta_T\|^2 - \frac{p}{D} \|\vbeta\|^2 \right|
  & = \left| \|\vbeta_{T^c}\|^2 - \left(1 - \frac{p}{D}\right) \|\vbeta\|^2 \right| .
\end{align*}
Observe that the finite population $(\beta_1^2,\dotsc,\beta_D^2)$ has mean $\tfrac1D \|\vbeta\|^2$, variance $\tfrac1D\sum_{j=1}^D \beta_j^4 - (\tfrac1D \sum_{j=1}^D \beta_j^2)^2 \leq \tfrac1D \|\vbeta\|^4 \mu^2 - (\tfrac1D \|\vbeta\|^2)^2 = \tfrac1D \|\vbeta\|^4 (\mu^2 - \tfrac1D)$, and range $\max_{j\in[D]} \beta_j^2 = \|\vbeta\|^2\mu^2$.
Therefore, Proposition~1.4 of \citep{bardenet2015concentration} and a union bound implies, with probability at least $1-2e^{-t}$,
\begin{align*}
  \left| \|\vbeta_T\|^2 - \frac{p}{D} \|\vbeta\|^2 \right|
  = \left| \|\vbeta_{T^c}\|^2 - \left(1 - \frac{p}{D}\right) \|\vbeta\|^2 \right|
  & \leq \|\vbeta\|^2 \left( \sqrt{2\left(\mu^2-\frac1D\right) \frac{pt}{D}} + \frac{2\mu^2 t}{3} \right) .
\end{align*}
If $p/D$ is more than $1/2$, then we can replace $p/D$ by $1-p/D$ on the right-hand side by analogously applying the previous argument to the random sample of cardinality $D-p$ that determines $\vbeta_{T^c}$.
\qed

\end{document}